\documentclass{article}

\PassOptionsToPackage{numbers, sort&compress}{natbib}

\usepackage[preprint]{neurips_2025}

\usepackage[utf8]{inputenc} 
\usepackage[T1]{fontenc}    
\usepackage{hyperref}       
\usepackage{url}            
\usepackage{booktabs}       
\usepackage{amsfonts}       
\usepackage{nicefrac}       
\usepackage{microtype}      
\usepackage{xcolor}         
\usepackage{hyperref}
\usepackage{multirow}
\usepackage{graphicx}
\usepackage{amsmath}
\usepackage{comment}
\usepackage{subcaption}
\usepackage{enumitem}

\title{From Virtual Games to Real-World Play}

\author{%
  Wenqiang Sun\footnotemark[1]~~$^{1,2}$,\quad Fangyun Wei\footnotemark[1]~~\footnotemark[2]~~$^{2}$,\quad Jinjing Zhao $^{3}$,\quad Xi Chen $^{2}$,\quad Zilong Chen$^{4}$, \\
    \textbf{Hongyang Zhang}$^{5}$, \quad \textbf{Jun Zhang}\footnotemark[2]~~$^{1}$,\quad \textbf{Yan Lu}$^{2}$ \\
  $^{1}$HKUST
~~$^2$Microsoft Research ~~$^3$University of Sydney \\
~~$^4$Tsinghua University ~~$^5$University of Waterloo\\
  \texttt{wsunap@connect.ust.hk, \{fawe,xichen6,yanlu\}@microsoft.com, } \\
\texttt{jzha0100@uni.sydney.edu.au, chenz122@mails.tsinghua.edu.cn, } \\
\texttt{hongyang.zhang@uwaterloo.ca, eejzhang@ust.hk} \\
}

\begin{document}

\maketitle

\renewcommand{\thefootnote}{\fnsymbol{footnote}}
\footnotetext[1]{Equal contribution.  \footnotemark[2]Corresponding author.}

\begin{figure}[h]
\centering
\includegraphics[width=\textwidth]{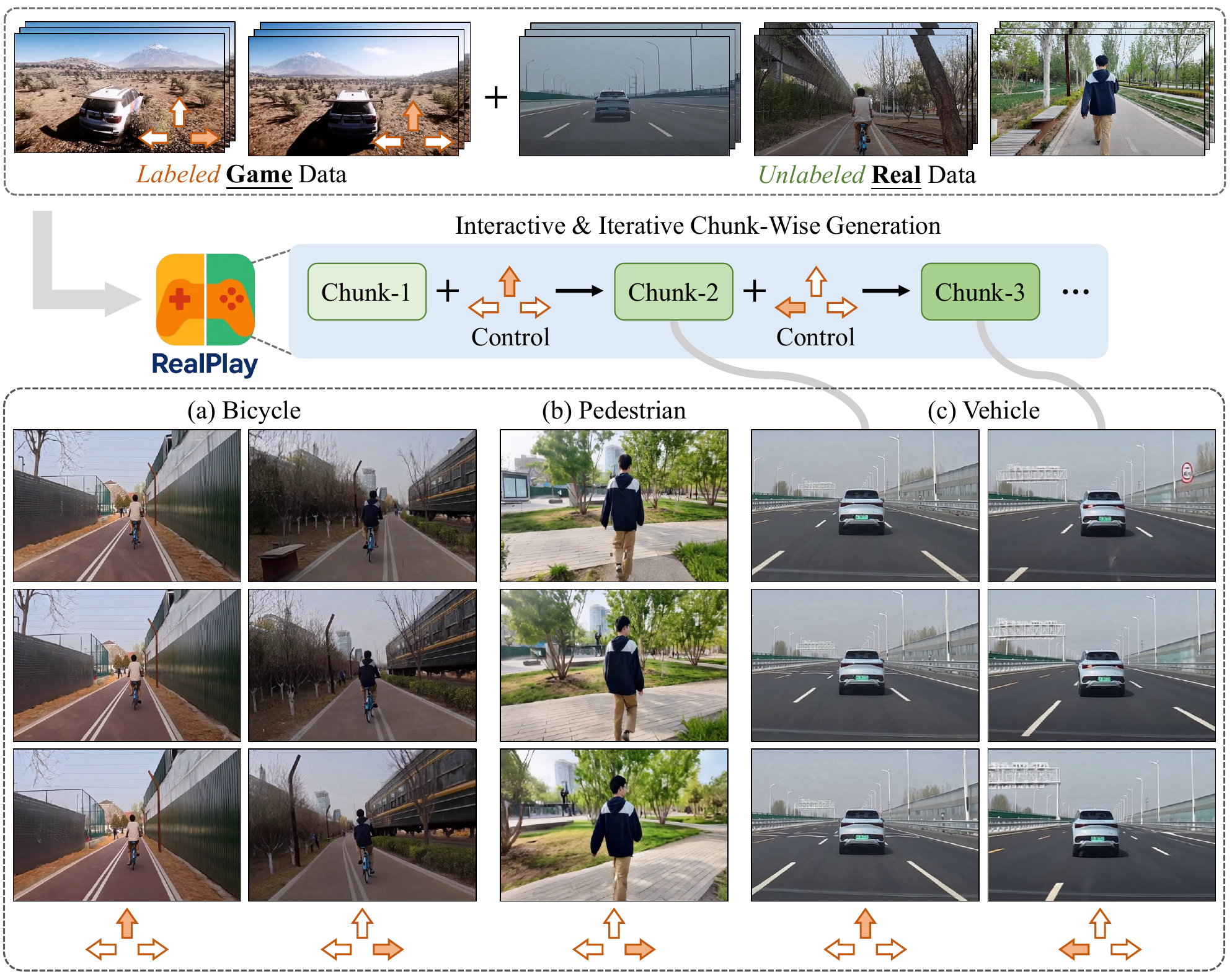}
\vspace{-6mm}
\caption{\textit{RealPlay} is a neural-network-driven real-world game engine with three key characteristics: (1) It supports iterative interaction—at each iteration, users observe the current visual scene, provide control signals, and receive control-accurate, temporally consistent, and realistic video chunks in response. (2) It eliminates the need for annotated real-world data while exhibiting strong control transfer capabilities, effectively mapping control signals (e.g., ``move forward'', ``turn left'' and ``turn right'') from the game environment to the real world. (3) It demonstrates entity transfer capabilities: although the labeled game data are sourced exclusively from the car racing game \textit{Forza Horizon 5}, RealPlay successfully generalizes these control signals to other real-world entities such as (a) bicycles and (b) pedestrians, beyond (c) vehicles. \textbf{Additional visualizations are provided in the appendix.}}
\label{fig:teaser}
\end{figure}

\vspace{-2mm}
\begin{abstract}
\vspace{-2mm}
We introduce RealPlay, a neural network-based real-world game engine that enables interactive video generation from user control signals. Unlike prior works focused on game-style visuals, RealPlay aims to produce photorealistic, temporally consistent video sequences that resemble real-world footage. It operates in an interactive loop: users observe a generated scene, issue a control command, and receive a short video chunk in response. To enable such realistic and responsive generation, we address key challenges including iterative chunk-wise prediction for low-latency feedback, temporal consistency across iterations, and accurate control response. RealPlay is trained on a combination of labeled game data and unlabeled real-world videos, without requiring real-world action annotations. Notably, we observe two forms of generalization: (1) control transfer—RealPlay effectively maps control signals from virtual to real-world scenarios; and (2) entity transfer—although training labels originate solely from a car racing game, RealPlay generalizes to control diverse real-world entities, including bicycles and pedestrians, beyond vehicles. Project page can be found: https://wenqsun.github.io/RealPlay/
\end{abstract}
\vspace{-3mm}
\section{Introduction}
\vspace{-2mm}
\label{sec:intro}

Recent works such as GameFactory~\cite{yu2025gamefactory} and GameNGen~\cite{valevski2024diffusion} demonstrate that neural networks can effectively model games and even function as game engines. These approaches typically adopt a two-stage pipeline: (1) collecting large-scale game data where each frame or chunk is annotated with control signals (e.g., ``move forward'' or ``turn left'' in a car racing game); and (2) training an interactive visual content generator—such as a video generation model—on the labeled data to predict future visual frames conditioned on the current visual context and the corresponding control signal.

Although neural networks can effectively model and replicate game environments, the upper bound of the visual quality they generate is ultimately constrained by the underlying game engine. While modern game engines—such as Unreal Engine 5—can produce highly realistic graphics, humans can still easily distinguish between game-rendered visuals and real-world footage. In other words, current game engines struggle to generate visuals that are indistinguishably close to reality or faithfully capture the complex physical laws of the real world. This motivates us to explore the feasibility of using neural networks to develop games whose visual outputs appear more realistic than those rendered by conventional game engines.

In this work, we present \emph{RealPlay}, an interactive video generation model that functions as a real-world game engine, built upon recent advances in diffusion-based video generation models~\cite{hong2022cogvideo, yang2024cogvideox, kong2024hunyuanvideo, zheng2024open1, peng2025open2, lin2024open_plan, wan2025}. Notably, RealPlay does not rely on labeled real-world data; instead, it leverages labeled game data and unlabeled real-world data, demonstrating strong control transfer capabilities from virtual environments to real-world scenarios. As illustrated in Figure~\ref{fig:teaser}, RealPlay enables users to engage in an interactive loop: they observe the current visual scene, provide control signals, and receive temporally consistent, photorealistic frames in response. The newly generated frames then serve as the observation for the next interaction step. RealPlay tackles four key challenges in interactive video generation:

    \noindent \textbf{Chunk-Wise Generation.} RealPlay is powered by a pre-trained image-to-video generator~\cite{yang2024cogvideox}, originally designed to produce long-horizon videos in a single forward pass. However, generating long clips at each iteration may result in noticeable delays, reducing the responsiveness of the interaction. To address this, we adapt the pre-trained video generator to produce shorter clips—consisting of only a few frames—at each iteration, enabling a more responsive user experience. 
    
    \noindent \textbf{Iterative Generation.} Existing video generation models are typically designed for one-shot image-to-video generation, which does not align with our use case where video chunks must be generated iteratively, with RealPlay receiving a new control signal at each iteration. To support this iterative generation process, we adapt a pre-trained video generator—originally conditioned on a single image to produce an entire video—into a chunk-wise generation framework, where each new segment is conditioned on the previously generated video chunk rather than just a static image.
    
    \noindent \textbf{Consistency Across Iterations.} During inference, RealPlay generates visual outputs conditioned on the current observation—which is itself generated in the previous iteration—and the user-provided control signals. In other words, each new video chunk is generated based on \textit{predicted observations}, rather than the \textit{ground-truth observations} used during training. This discrepancy introduces a distribution gap between training and inference, often resulting in accumulated artifacts, visual drift, or temporal inconsistencies across iterations. To mitigate this issue, we adopt the Diffusion Forcing~\cite{chen2024diffusion} strategy, which has been shown effective in narrowing this gap. Specifically, during training, we introduce noise to the conditional chunk, encouraging the model to better handle imperfect conditions during inference. 
    
    \noindent \textbf{Precise Control.} Annotating game data is relatively easy and highly scalable, as control signals can be automatically recorded during gameplay.  In contrast, labeling real-world data is time-consuming, ambiguous, and often requires manual efforts. RealPlay explores a mixed training paradigm that combines labeled game data with unlabeled real-world data. Our findings show that this approach enables RealPlay to learn effective control strategies in the game domain and successfully transfer them to real-world scenarios—demonstrating promising control transfer capability without requiring explicit real-world annotations. Interestingly, although our labeled game data come solely from the car racing game \textit{Forza Horizon 5}, the control signals (i.e., ``move forward'', ``turn left'', and ``turn right'') can be effectively transferred to control real-world entities beyond vehicles—such as bicycles and pedestrians—as illustrated in Figure~\ref{fig:teaser}.

Our experimental results demonstrate that RealPlay is capable of generating visually compelling, control-accurate, and realistic video sequences, while supporting iterative user interaction. RealPlay represents an initial step toward validating the feasibility of using neural networks as real-world game engines—paving the way for a new paradigm where interactive, high-fidelity simulations are driven purely by data and learned dynamics, rather than traditional graphics engines.

\vspace{-3mm}
\section{Related Work}
\label{sec:related_work}
\vspace{-2mm}

\textbf{Video Diffusion Models.} The field of video generation has advanced rapidly with the emergence of diffusion models~\cite{ho2022video, chen2023videocrafter1, chen2024videocrafter2}, with modern architectures demonstrating exceptional temporal coherence and physical plausibility~\cite{kong2024hunyuanvideo, yang2024cogvideox, wan2025, liu2024reconx, sun2024dimensionx}. These models support multi-modal conditioning—ranging from text-to-video synthesis~\cite{singer2022make, hong2022cogvideo, yang2024cogvideox} to image animation~\cite{siarohin2019first, guo2023animatediff, xing2024dynamicrafter}—while maintaining stable dynamics over extended sequences. Beyond creative applications, their ability to capture realistic dynamics positions them as promising candidates for world simulators~\cite{oasis2024, bruce2024genie, parkerholder2024genie2, feng2024matrix, ye2024latent}, with early demonstrations in tasks such as game engines and embodied planning.

Within this landscape, bidirectional diffusion models have gained prominence by employing global spatio-temporal attention to capture inter-frame dependencies, enabling the generation of high-fidelity videos with strong temporal coherence~\cite{kong2024hunyuanvideo, zheng2024open1, lin2024open_plan, peng2025open2, ma2025step, wan2025}. However, their quadratic computational complexity—stemming from full-sequence attention—limits them to generating only short-horizon videos. To overcome this limitation, autoregressive diffusion frameworks have been introduced, generating videos iteratively by conditioning on previously generated frames~\cite{kim2024fifo, jin2024pyramidal, alonso2024diffusion, valevski2024diffusion, gao2024vid, henschel2024streamingt2v, wu2022nuwa}. While early variants successfully extend sequence lengths, they often suffer from performance degradation over time, leading to diminished visual quality and temporal consistency. To address this, diffusion forcing techniques are proposed, introducing flexible temporal conditioning mechanisms that significantly mitigate these issues~\cite{chen2024diffusion, song2025history}. Building upon this foundation, recent autoregressive methods have further improved long-term coherence and visual fidelity, enabling efficient and stable generation of significantly longer videos~\cite{yin2024slow, gu2025long, wu2023ar, xie2024progressive, magi1, chen2025skyreels}.

\textbf{World Models.} Building on the generative strengths of video diffusion models, recent research has explored their potential as interactive world models capable of simulating complex, controllable environments~\cite{alonso2024diffusion, valevski2024diffusion, che2024gamegen, yang2024playable, oasis2024, bruce2024genie, xiao2025worldmem}. These efforts move beyond passive video generation toward agent-environment interaction, using diffusion-based architectures to synthesize responsive and dynamic virtual worlds. Foundational works~\cite{feng2024matrix, yu2025gamefactory, parkerholder2024genie2, zhang2025matrixgame} highlight the feasibility of generating playable environments from diverse inputs—such as images, text, or user actions—thus enabling real-time interaction and embodied planning. However, while many of these models produce visually rich simulations, they often suffer from limited generalization due to overfitting on specific virtual datasets. In contrast, RealPlay targets real-world scenarios, enabling photorealistic and controllable video generation from user input, while demonstrating strong generalization across domains and entities.

\vspace{-3mm}
\section{Method}
\label{sec:method}
\vspace{-2mm}

\begin{figure}[t]
\centering
\includegraphics[width=\textwidth]{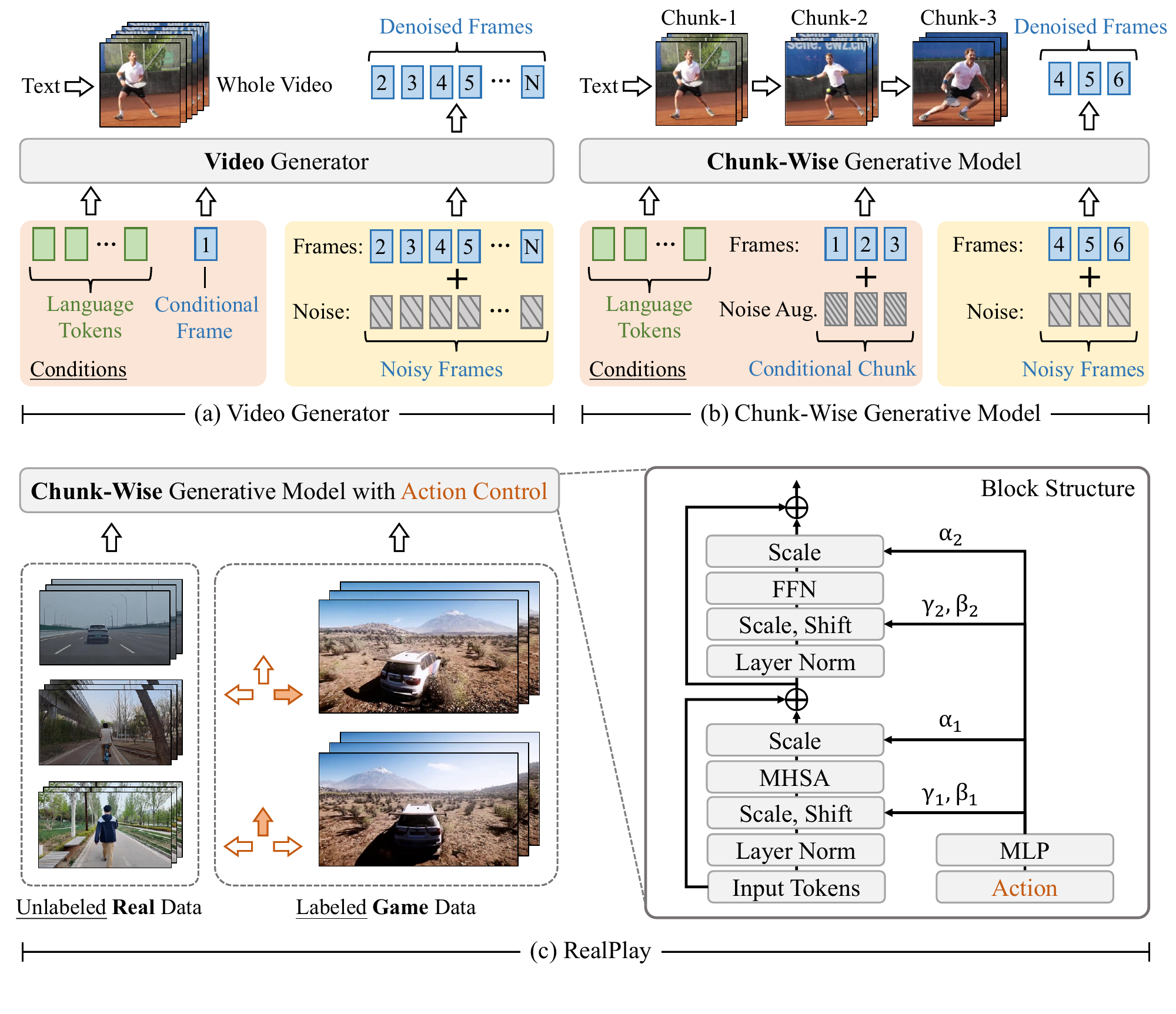}
\vspace{-6.5mm}
\caption{RealPlay involves a two-stage training process. \textit{Stage-1}: We adapt a pre-trained image-to-video generator (Figure~(a))—which generates an entire video in a single pass conditioned on a single frame—into a chunk-wise generation model (Figure~(b)), which generates video chunks iteratively, conditioned on the previously generated chunk. This adaptation includes several key modifications detailed in Section~\ref{sec:chunk-wise}. \textit{Stage-2}: RealPlay (Figure~(c)) is trained on a combination of a labeled game dataset and an unlabeled real-world dataset, enabling action transfer from controlling a car in the game environment to manipulating various entities in the real world. This is achieved by modifying the chunk-wise generation model to incorporate action control through an adaptive LayerNorm mechanism. In all figures, ``frames'' refer to frame latents encoded by the video VAE encoder from CogVideoX~\cite{yang2024cogvideox}. For clarity, we omit the details of injecting noise timestep embeddings. } 
\label{fig:overview}
\vspace{-3mm}
\end{figure}

\noindent \textbf{Problem Formulation.} The objective is to train \textit{RealPlay} using a labeled car racing dataset, in which user actions (i.e., ``\texttt{move forward}'', ``\texttt{turn left}'', and ``\texttt{turn right}'') are recorded at a fixed frequency, alongside an unlabeled dataset consisting of real-world videos that capture featuring vehicles in motion, bicycles being ridden, and pedestrians walking, as shown in Figure~\ref{fig:teaser}. Once trained, RealPlay enables control transfer from car-based actions in the game environment to real-world entities including vehicles, bicycles, and pedestrians. It supports interactive and iterative video generation: at each iteration, users observe the scene generated in the previous step and issue new control commands. RealPlay then generates a new, temporally consistent video chunk that reflects the user’s input.

\noindent \textbf{Training Data.} Formally, let $\mathcal{G}=\{G_i\}_{i=1}^{M_1}$ denote the \textit{labeled} game dataset comprising $M_1$ training samples. Each sample $G_i=\{(\boldsymbol{C}_k,\boldsymbol{a}_k)\}_{k=1}^{K}$ consists of $K$ pairs of video chunks and corresponding action commands. An inherent temporal association exists between adjacent pairs: the $k$-th video chunk $\boldsymbol{C}_k$ combined with the action $\boldsymbol{a}_k$ leads to the $(k+1)$-th video chunk $\boldsymbol{C}_{k+1}$. Each action $\boldsymbol{a}_k$ is represented as a 3-dimensional one-hot vector, corresponding to the discrete set \{\texttt{move forward}, \texttt{turn left}, \texttt{turn right}\}. Let $\mathcal{R}=\{R_i\}_{i=1}^{M_2}$ represent the \textit{unlabeled} real-world dataset including $M_2$ samples. Each sample $R_i=\{\boldsymbol{C}_k\}_{k=1}^{K}$ contains a sequence of $K$ video chunks, where adjacent chunks exhibit temporal consistency. Each sequence captures real-world motion patterns involving vehicles, bicycles, or pedestrians, each appearing exclusively within a given sequence.

\vspace{-0.5mm}
\noindent \textbf{Overview.} Figure~\ref{fig:overview} presents an overview of RealPlay, a two-stage training pipeline: 
\vspace{-2mm}
\begin{enumerate}[leftmargin=1em]
    \item In stage 1, we adapt a pre-trained image-to-video generator—originally designed to generate long videos conditioned on clean, ground-truth images—into a chunk-wise generative model capable of generating video segments conditioned on noisy, previously generated chunks. Details are provided in Section~\ref{sec:chunk-wise}. \vspace{-0.5mm}
    \item In stage 2, we fine-tune this chunk-wise model using both the \textit{labeled} game dataset $\mathcal{G}$ and the \textit{unlabeled} real-world dataset $\mathcal{R}$, ultimately training RealPlay to perform chunk-wise generation while enabling control transfer from the game to the real world, as detailed in Section~\ref{sec:mixed-training}.
\end{enumerate}

\subsection{Chunk-Wise Generation}
\vspace{-2mm}
\label{sec:chunk-wise}
\textbf{Image-to-Video Generator.} We begin by adapting a pre-trained image-to-video generator into a chunk-wise generative model. Specifically, we leverage CogVideoX, a large-scale model capable of generating videos with a duration of 49 frames. CogVideoX~\cite{hong2022cogvideo} employs T5~\cite{2020t5} as its text encoder, which encodes a language instruction into a fixed-length token sequence denoted as $\mathcal{T}$. It also includes a video VAE encoder with a downsampling rate of 4, which encodes each video into a sequence of latent representations $\mathcal{F}=\{\boldsymbol{F}_i\}_{i=1}^{N}$, where $N$ denotes the number of video latents (with $N=13$ in the case of CogVideoX). Due to the causal design of the video VAE encoder, the latent $\boldsymbol{F}_1$ captures information solely from the first frame. The remaining latents $\{\boldsymbol{F}_i\}_{i=2}^{N}$ correspond to the future frames to be generated. As illustrated in Figure~\ref{fig:overview}(a), CogVideoX's backbone is a DiT~\cite{peebles2023scalable}, which takes as input the concatenation of the language tokens $\mathcal{T}$, the first frame latent $\boldsymbol{F}_1$, and a sequence of noisy future frame latents. All frame latents are flattened in raster scan order. Each token, regardless of its modality, attends to every other token. The model is trained using a DDPM-style~\cite{ho2020denoising} diffusion loss applied over $\{\boldsymbol{F}_i\}_{i=2}^{N}$. Further details are available in the original CogVideoX paper.

\vspace{-0.5mm}
\noindent\textbf{Adaptation to Chunk-Wise Generation.} As illustrated in Figure~\ref{fig:overview}(b), our adaptation of the original model to enable chunk-wise generation involves four key modifications: \vspace{-1.5mm}
\begin{enumerate}[leftmargin=1em]
    \item \textbf{Chunk-Level Conditioning.} We replace the original first-frame condition with a chunk-level condition, enabling the model to condition on a previously generated video chunk rather than a single frame. \vspace{-0.5mm}
    \item \textbf{Masking Strategy.} The attention mask is tailored for chunk-wise generation: latents in the conditioning chunk can attend only to the language tokens and themselves, while latents of future frames can attend to all tokens (language tokens, conditioning chunk latents, and themselves). \vspace{-0.5mm}
    \item \textbf{Temporal Resolution Adjustment.} We reduce the number of future frame latents to be generated from 13 (covering 49 frames in CogVideoX) to 4 (covering 16 frames), enabling finer-grained iterative generation with lower latency. \vspace{-0.5mm}
    \item \textbf{Noise Augmentation.} During training, the model is conditioned on ground-truth video chunks, whereas at inference time, it relies on its own generated outputs, which may deviate from the true distribution. To address this mismatch, we inject noise into the conditioning inputs during training—a strategy shown to effectively improve robustness~\cite{chen2024diffusion}. At inference, the model directly uses previously generated latents as conditioning inputs for the current iteration. \vspace{-0.5mm}
\end{enumerate}
\vspace{-0.5mm}
All other configurations remain consistent with those of CogVideoX~\cite{yang2024cogvideox}, including the use of its backbone, language encoder, video VAE encoder, DDIM-style~\cite{song2020denoising} inference, and the strategy for applying multi-modal positional embeddings. The training in this stage is performed on a general-domain video dataset. Specifically, we select 100K high-quality videos from the OpenViD~\cite{nan2024openvid} dataset to serve as the training corpus.

\vspace{-2mm}
\subsection{From Game Environment to Real World}
\vspace{-1.5mm}
\label{sec:mixed-training}
RealPlay is trained on a combination of a labeled game dataset $\mathcal{G}$ and an unlabeled real-world dataset $\mathcal{R}$, enabling control transfer from operating a car in the game environment to controlling vehicles, bicycles, and pedestrians in real-world scenarios. We fine-tune the chunk-wise generation model introduced in Section~\ref{sec:chunk-wise} on both $\mathcal{G}$ and $\mathcal{R}$, introducing a control module as the only architectural modification to enable action-conditioned chunk-wise generation.

Specifically, given a sample $G_i=\{(\boldsymbol{C}_k,\boldsymbol{a}_k)\}_{k=1}^{K}$ from the labeled dataset $\mathcal{G}$—where each $\boldsymbol{C}_k$ denotes the $k$-th video chunk and $\boldsymbol{a}_k$ is a 3-dimensional one-hot vector indicating the action (from the discrete set \{\texttt{move forward}, \texttt{turn left}, \texttt{turn right}\}) that transitions $\boldsymbol{C}_k$ to $\boldsymbol{C}_{k+1}$—the control module learns to condition the generation process on the specified action. 

As illustrated in Figure~\ref{fig:overview}(c), RealPlay extends the chunk-wise generation architecture described in Section~\ref{sec:chunk-wise} by introducing an additional input: the action $\boldsymbol{a}_k$. This action is injected via an adaptive LayerNorm mechanism. Specifically, $\boldsymbol{a}_k$ is first projected into a 512-D feature vector using an MLP. This vector is then added to the noise timestep embedding, and the resulting feature is passed through a linear layer to produce two sets of modulation parameters: $\{\alpha_1, \gamma_1, \beta_1\}$ and $\{\alpha_2, \gamma_2, \beta_2\}$, where $\alpha$ and $\gamma$ are scale factors and $\beta$ is a shift bias. These parameters modulate the network activations before and after the attention and feed-forward layers, respectively, enabling action-aware generation.

For a sample $R_i = \{\boldsymbol{C}_k\}_{k=1}^{K}$ from the real-world dataset $\mathcal{R}$, action annotations are unavailable, so we only observe the visual transition from $\boldsymbol{C}_k$ to $\boldsymbol{C}_{k+1}$. During training, we represent the absence of explicit action supervision by using an all-zero vector $\boldsymbol{0}$ as the action input.

In summary, during training, we adopt the formulation $\boldsymbol{C}_k + \boldsymbol{a}_k\rightarrow \boldsymbol{C}_{k+1}$ for labeled game samples, while we use $\boldsymbol{C}_k + \boldsymbol{0}\rightarrow \boldsymbol{C}_{k+1}$ for unlabeled real-world samples. During inference, given a previously generated video chunk $\boldsymbol{C}_k$ depicting a real-world scene (e.g., a moving car, a riding bicycle, or a walking pedestrian), we apply an action command originating from the game environment to generate the next video chunk $\boldsymbol{C}_{k+1}$, which faithfully follows the specified action.

\vspace{-0.5mm}
We now provide an intuitive explanation for the emergence of \textit{game-to-real-world transfer capabilities}—from controlling a car in a game environment to controlling various entities such as vehicles, bicycles, and pedestrians in the real world—through the lenses of classifier-free guidance (CFG)~\cite{ho2022classifier} and network generalization. CFG is a technique used in conditional generation models that in the training stage, CFG randomly drops conditioning inputs of a training sample with a pre-defined probability, while during inference it steers outputs toward a desired condition by interpolating between conditional and unconditional predictions. 

\vspace{-0.5mm}
We begin by introducing RealPlay trained solely on the labeled game dataset $\mathcal{G}$, with the incorporation of CFG. Under this setting, each sample from $\mathcal{G}$ has a predefined probability of being used with or without its action condition during training. This mechanism is effectively equivalent to partitioning $\mathcal{G}$ into two disjoint subsets: $\mathcal{G}_L$, which retains action labels, and $\mathcal{G}_U$, which omits them. RealPlay is jointly trained on both subsets, resulting in controllable generation performance that closely matches the variant trained on $\mathcal{G}$ with conventional CFG. 

In our actual scenario, we use the labeled game dataset $\mathcal{G}$ as the counterpart to $\mathcal{G}_L$, and the unlabeled real-world dataset $\mathcal{R}$ as the counterpart to $\mathcal{G}_U$. Although $\mathcal{G}$ and $\mathcal{R}$ originate from different distributions, they share important commonalities: (1) both datasets capture entity motion—while $\mathcal{G}$ primarily depicts cars in motion, $\mathcal{R}$ includes additional entities such as bicycles and pedestrians; and (2) the game data used in our setup are derived from Forza Horizon 5, a photorealistic AAA game with high visual realism, which significantly narrows the domain gap between synthetic and real-world data. Combined with the strong generalization capabilities of the large-scale pre-trained video generator, RealPlay enables control transfer from operating a car in the game environment to manipulating diverse entities in the real world.

\vspace{-3mm}
\section{Experiment}
\vspace{-3mm}
\label{sec:Exp}
\noindent \textbf{Implementation Details.} We use CogVideoX-5B~\cite{yang2024cogvideox} as our pre-trained model. First, we fine-tune CogVideoX-5B for chunk-wise generation, as described in Section~\ref{sec:chunk-wise}, using 100K samples from the OpenViD~\cite{nan2024openvid} dataset over 10K iterations on 8$\times$A100 GPUs. We then further fine-tune the model for 15K iterations, also on 8$\times$A100 GPUs, following the mixed training strategy outlined in Section~\ref{sec:mixed-training}. This stage uses a combined dataset comprising a labeled game dataset from The Matrix~\cite{feng2024matrix}—which contains 80K samples from the car racing game Forza Horizon 5—and our own collected unlabeled real-world dataset, consisting of 18K samples each for vehicles, bicycles, and pedestrians. Each game sample is a 32-frame video clip annotated with an action label that controls the transition from the first 16 frames to the next 16. In contrast, each real-world sample is an unannotated 32-frame clip, capturing unconstrained motion of entities. Additional training details are provided in the appendix.

\vspace{-0.5mm}
\noindent \textbf{Evaluation.} Our evaluation covers three key aspects:
\vspace{-0.5mm}
\begin{enumerate}[leftmargin=1em]
    \item \textbf{Visual Quality.} We adopt four metrics proposed by VBench~\cite{huang2024vbench}—motion consistency, aesthetic appeal, imaging quality, and scene dynamics—to quantitatively evaluate the visual quality of the generated video chunks.\vspace{-0.5mm}
    \item \textbf{Control Effectiveness.} We report the success rate based on human evaluation, where evaluators judge whether the generated video accurately reflects the intended control command.\vspace{-0.5mm}
    \item \textbf{Comprehensive Human Evaluation.} To holistically evaluate model performance, we compute an Elo score for each model using 500 pairwise comparisons. In each comparison, two video outputs—generated from the same initial observation and control condition by different models—are displayed side by side. Human evaluators are asked to select the video that appears more realistic and better aligned with the intended control. Elo scores are updated using a standard rating adjustment scheme, where the winner gains points from the loser. All models are initialized with an Elo score of 1000.\vspace{-0.5mm}
\end{enumerate}

\vspace{-1mm}
Unless otherwise specified, we iteratively generate 3 chunks using 3 control commands, each randomly selected from the set \{\texttt{move forward}, \texttt{turn left}, \texttt{turn right}\}, for comparisons with other models and for ablation studies. We evaluate the performance on vehicles, bicycles, and pedestrians that were observed during training.

\begin{table}[!t]
\centering
\small
\caption{Comparison of \textit{RealPlay variants} with: (1) \textit{Single-forward-pass models}, which generate an entire video sequence in a single forward pass; (2) \textit{Chunk-wise generation models}, where CogVideoX-5B is first adapted to a chunk-wise generator using the approach described in Section~\ref{sec:chunk-wise}, and then fine-tuned on either manually labeled real-world data or pseudo-labeled data generated by LAPA~\cite{ye2024latent}. RealPlay-AdaLN: uses Adaptive LayerNorm to fuse action signals. RealPlay-Text: actions are expressed in text prompt. `L'' and ``U'' denote labeled data and unlabeled data, respectively.} 
\setlength{\tabcolsep}{3.5pt}
\begin{tabular}{l|cc|cccc|c|c}
\toprule
\multirow{2}{*}{Method} & Real & Game & \multicolumn{4}{c|}{Visual Quality} & Control & \multirow{2}{*}{Elo}\\ 
  & Data & Data & Motion & Aesthetic & Imaging  & Dynamic & Rate & \\
\midrule
\multicolumn{9}{l}{\textit{Single-Forward-Pass Models:}} \\
~CogVideoX-5B~\cite{yang2024cogvideox} & \multirow{4}{*}{-}& \multirow{4}{*}{-}  & 98.3 & 47.4 & 71.3 & 49.4 & 33.9 & 1025 \\
~Hunyuan-720P~\cite{kong2024hunyuanvideo} & &  & 99.3 & 49.1 & 91.9 & 52.8 & 31.7 & 1102\\
~Wan-2.1~\cite{wan2025} & &  & 97.9 & 47.5 & 72.9 & 61.1 & 32.2 & 929\\
~OpenSora-2.0~\cite{peng2025open2} & &  & 99.2 & 46.9 & 69.5 & 100.0 & 26.7 & 795\\
\midrule
\multicolumn{9}{l}{\textit{Chunk-Wise Generation Models:}} \\
~CogVideoX-5B (Human)& L& -  & 97.9 & 46.7 & 69.7 & 98.5 & 58.9 & 1045 \\
~CogVideoX-5B (LAPA~\cite{ye2024latent}) & U & - & 98.7 & 44.5 & 64.3 & 62.3 & 36.2 & 779\\
\midrule
\multicolumn{9}{l}{\textit{RealPlay Variants:}} \\
 ~RealPlay-Text & U & L & 97.6 & 46.3 & 69.0 & 99.4 & 69.9 & 1143 \\
~RealPlay-AdaLN (Default) &  U &  L & 97.5 & 47.8 & 68.9 & 100.0 & 90.0 & \textbf{1184} \\
\bottomrule
\end{tabular}
\vspace{-4mm}
\label{tab:main}
\end{table}

\vspace{-2mm}
\subsection{Main Results}
\vspace{-2mm}
In Table~\ref{tab:main}, we compare our RealPlay model with several baselines, which can be categorized into four classes. All models receive the same initial visual input, but differ in their approach to action control and the nature of their training data: \vspace{-2.5mm}
\begin{enumerate}[leftmargin=1em]
    \item \textbf{Single-Forward-Pass Models.} We directly evaluate four state-of-the-art pre-trained (text,image)-to-video models—CogVideoX-5B~\cite{yang2024cogvideox}, Hunyuan-720P~\cite{kong2024hunyuanvideo}, Wan-2.1~\cite{wan2025}, and OpenSora-2.0~\cite{peng2025open2}—without any fine-tuning. The text prompt is ``Control the [\texttt{Entity}] to first [\texttt{Action-1}], then [\texttt{Action-2}], and finally 
 [\texttt{Action-3}]'', where \texttt{Entity} $\in$ \{\texttt{vehicle}, \texttt{bicycle}, \texttt{pedestrian}\}, and \texttt{Action-1,2,3} $\in$ \{\texttt{move forward}, \texttt{turn left}, \texttt{turn right}\}.\vspace{-1mm}
 \item \textbf{Chunk-Wise Generation Models Trained on Labeled Real-World Data.} We adapt the original CogVideoX-5B~\cite{yang2024cogvideox} into a chunk-wise generation model using the method described in Section~\ref{sec:chunk-wise}, and fine-tune it on our collected real-world dataset using a selective annotation strategy. Specifically, we manually label only clear samples and exclude ambiguous ones to ensure annotation quality, ultimately selecting approximately 20\% of the total data. Each training sample is a short video clip consisting of two temporally continuous chunks, denoted as $\boldsymbol{C}_1$ and $\boldsymbol{C}_2$. A sample is considered clear if, given $\boldsymbol{C}_1$, human annotators can confidently determine whether a single action $\boldsymbol{a}$ leads to $\boldsymbol{C}_2$ (e.g., moving forward throughout $\boldsymbol{C}_2$). In contrast, ambiguous samples contain multiple actions within $\boldsymbol{C}_2$, making it difficult to assign a single action label. Using this filtering strategy, we construct a training set in which we have reliable transitions of the form $\boldsymbol{C}_1 + \boldsymbol{a} \rightarrow \boldsymbol{C}_2$. \vspace{-1mm}
 \item \textbf{Chunk-Wise Generation Models Trained on Pseudo-Labeled Real-World Data.} 
This model is identical to the one described in point 2, except for the annotation strategy applied to the real-world data. Instead of relying on manual labels, we use LAPA~\cite{ye2024latent} to generate pseudo-action labels $\boldsymbol{a}'$. Details of the LAPA method are provided in the appendix. Using these pseudo labels, we construct a training set with transitions of the form $\boldsymbol{C}_1 + \boldsymbol{a}' \rightarrow \boldsymbol{C}_2$. \vspace{-0.5mm}
 \item \textbf{RealPlay.} We evaluate two variants of RealPlay: (1) \textit{RealPlay-AdaLN}: the default version, which uses Adaptive LayerNorm to fuse action signals. It is trained on both labeled game data and unlabeled real-world data. (2) \textit{RealPlay-Text}: trained with the same data as RealPlay-AdaLN, but action signals are reflected in the form of text instructions. Specifically, we use the template: ``Control the car to [\texttt{Action}]'' where \texttt{Action} $\in$ \{\texttt{move forward}, \texttt{turn left}, \texttt{turn right}\}. \vspace{-0.5mm}
\end{enumerate}

\begin{table}[!t]
\centering
\small
\caption{Per-entity evaluation shows that real-world entities with larger motion amplitudes achieve higher control success rates, as their more distinctive motion patterns make it easier for the network to transfer control from the game environment to the real world.}
\begin{tabular}{l|cccc|c|c}
\toprule
\multirow{2}{*}{Entity} & \multicolumn{4}{c|}{Visual Quality} & Control & Motion  \\ 
 & Motion & Aesthetic & Imaging & Dynamic & Rate&  Amplitude \\ 
   \midrule
       In-Game Car & 97.8 & 52.5 & 70.8 & 100.0 & 100.0 & 3.3 \\
    \midrule
    Vehicle & 97.1 & 52.6 & 71.0 & 100.0 & 83.3 & 1.9\\
    Bicycle & 97.1 & 51.2 & 66.8 & 100.0 & 91.7 & 2.8\\
    Pedestrian & 98.4 & 39.7 & 68.8 & 100.0 & 95.0 & 5.7 \\
\bottomrule
\end{tabular}
\vspace{-6mm}
\label{tab:category}
\end{table}

\begin{table}[!t]
\centering
\small
\caption{Cross-entity training significantly improves control success for individual entities. The ablation study is conducted on the \textit{bicycle} entity.}
\begin{tabular}{c|ccc|cccc|c}
\toprule
Game & \multicolumn{3}{c|}{Real-World Data} & \multicolumn{4}{c|}{Visual Quality} & Control\\ 
Data & Bicycle & Vehicle & Pedestrian & Motion & Aesthetic & Imaging  & Dynamic & Rate\\ 
   \midrule
  \checkmark & & & & 96.1 & 50.5 & 59.1 & 100.0 & 0.0 \\
  \checkmark & \checkmark & & & 97.1 & 50.3 & 67.0 & 100.0 & 72.5 \\
  \checkmark & \checkmark &\checkmark & & 96.7 & 52.7 & 66.7 & 100.0 & 78.4 \\
 \checkmark & \checkmark &\checkmark & \checkmark & 97.1 & 51.2 & 66.8 & 100.0 & 91.7 \\
\bottomrule
\end{tabular}
\vspace{-4mm}
\label{tab:training-data}
\end{table}

\vspace{-1.5mm}
From the results presented in Table~\ref{tab:main}, we draw several key conclusions: (1) All single-forward-pass models exhibit some level of control capability through text prompts. However, their control success rates remain low, ranging from 26.7\% to 33.9\%. We choose CogVideoX-5B~\cite{hong2022cogvideo} as our pre-trained model due to its highest zero-shot control performance and acceptable visual quality. Additionally, we observe that models such as CogVideoX-5B~\cite{yang2024cogvideox}, Hunyuan-720P~\cite{kong2024hunyuanvideo}, and Wan-2.1~\cite{wan2025} sometimes generate static videos, which contributes to lower dynamic scores. (2) The two chunk-wise generation models, fine-tuned on a human-labeled dataset and the LAPA~\cite{ye2024latent} pseudo-labeled dataset respectively, both show improvements in control success rate. However, the overall success rates remain low. This is primarily due to two factors: the limited size of the human-labeled dataset, which is insufficient to effectively train a large network, and the noise present in the pseudo labels, which hampers learning in the LAPA-fine-tuned model. (3) Both RealPlay variants significantly improve the control success rate. Our default implementation, which uses adaptive LayerNorm for action fusion, achieves a success rate of 90\%. Additionally, neither variant generates static videos; instead, they produce more visually appealing results than the original CogVideoX-5B, while introducing two key enhancements: chunk-wise generation and control over real-world entities in the video.

\vspace{-3mm}
\subsection{Analysis}
\vspace{-2mm}
\noindent\textbf{Larger Motion Amplitudes Yield Higher Control Rates.} RealPlay is trained using a labeled game dataset (car racing) and an unlabeled real-world dataset containing vehicles, bicycles, and pedestrians. In Table~\ref{tab:category}, we report the visual quality and control rate for each entity. A key observation is that control rates vary across real-world entities, ranging from 83.3\% to 95.0\%. We attribute this variation to differences in motion amplitude across real-world entities. Specifically, real-world pedestrians typically exhibit larger motion amplitudes than bicycles and vehicles—especially during dynamic movements such as turning—whereas vehicles tend to move more smoothly and slowly. Our study reveals that a larger motion amplitude leads to a higher control rate, as it results in more distinctive motion patterns—making it easier for the network to transfer control from the game environment to the real world. Details on computing motion amplitudes are provided in the appendix.

\begin{figure}[t]
    \centering
    \begin{minipage}[b]{0.48\textwidth}
        \includegraphics[width=0.98\textwidth]{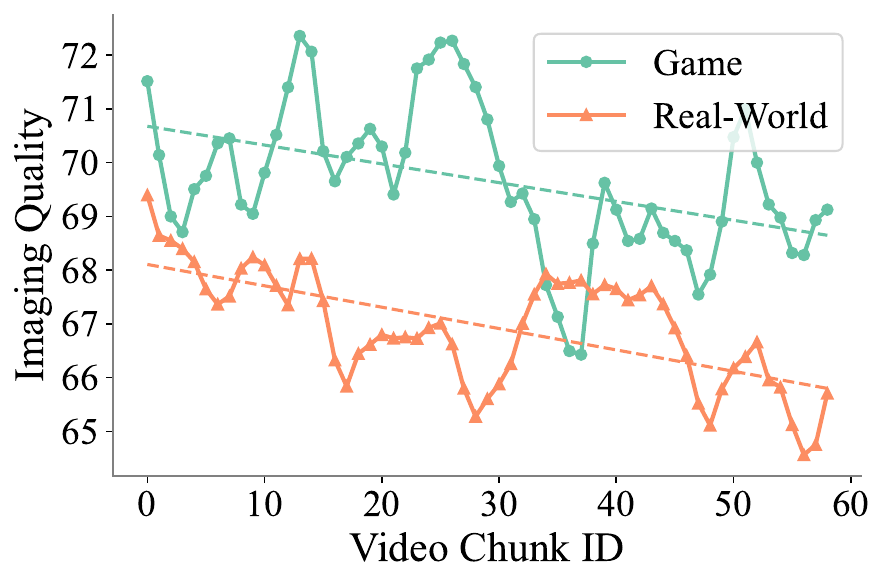}
        \vspace{-3mm}
        \caption{Visual quality degrades in both game and real-world settings, but the image quality when controlling a game entity consistently remains higher than that of a real-world entity (e.g., the bicycle in this study), highlighting the greater challenge of modeling real-world entities.}
        \vspace{-5mm}
         \label{fig:real_game}
    \end{minipage}
    \hfill
    \begin{minipage}[b]{0.48\textwidth}
        \includegraphics[width=0.98\textwidth]{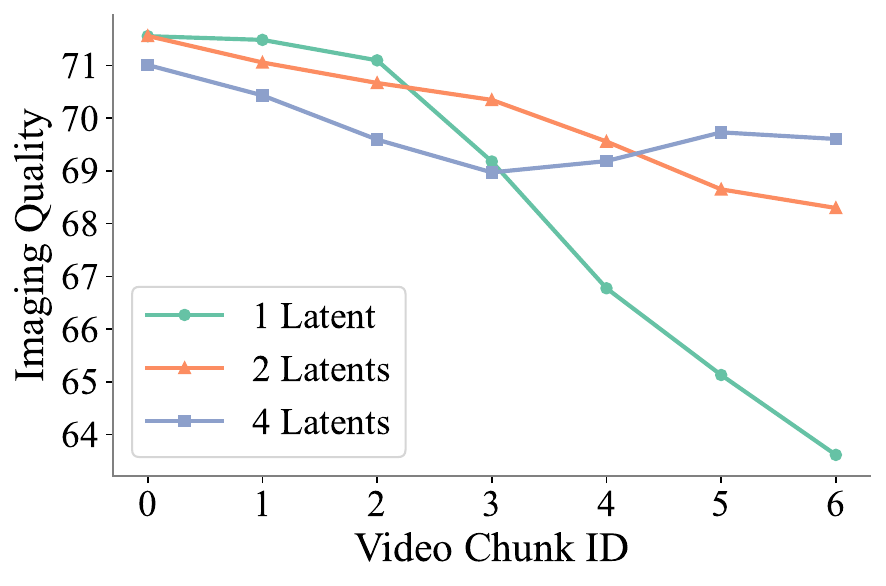}
        \vspace{-3mm}
        \caption{Reducing the number of video latents per chunk leads to visual quality degradation, as the pre-trained video generator—originally optimized for long-horizon generation—loses temporal coherence and consistency when adapted to extremely short-horizon outputs (e.g., 1 latent).}
        \vspace{-5mm}
        \label{fig:chunk_x}
    \end{minipage}
\end{figure}

\begin{figure}[t]
    \centering
    \begin{minipage}[b]{0.48\textwidth}
        \includegraphics[width=0.98\textwidth]{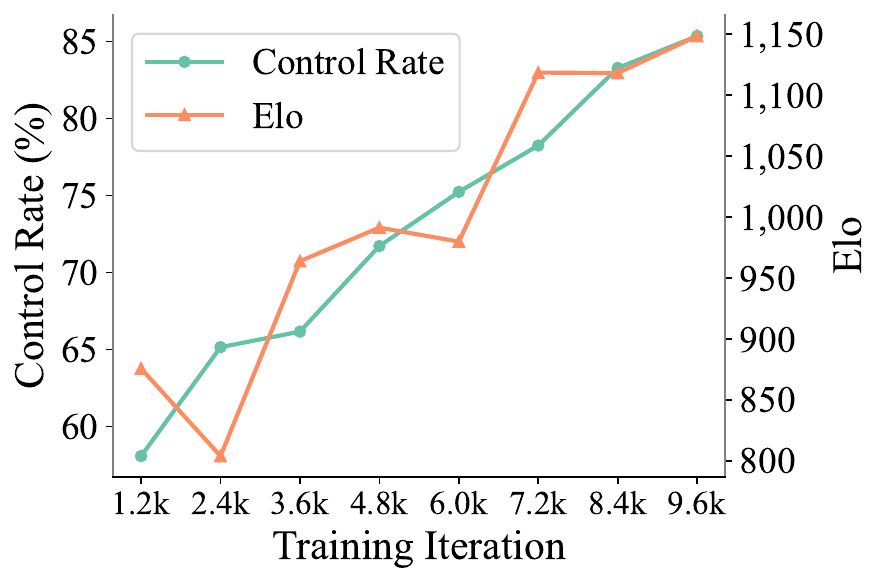}
        \vspace{-3mm}
        \caption{Both the control success rate and Elo scores steadily improve as training progresses. The evaluation is performed on the bicycle entity.}
        \vspace{-6mm}
        \label{fig:iteration}
    \end{minipage}
    \hfill
     \begin{minipage}[b]{0.48\textwidth}
        \includegraphics[width=0.98\textwidth]{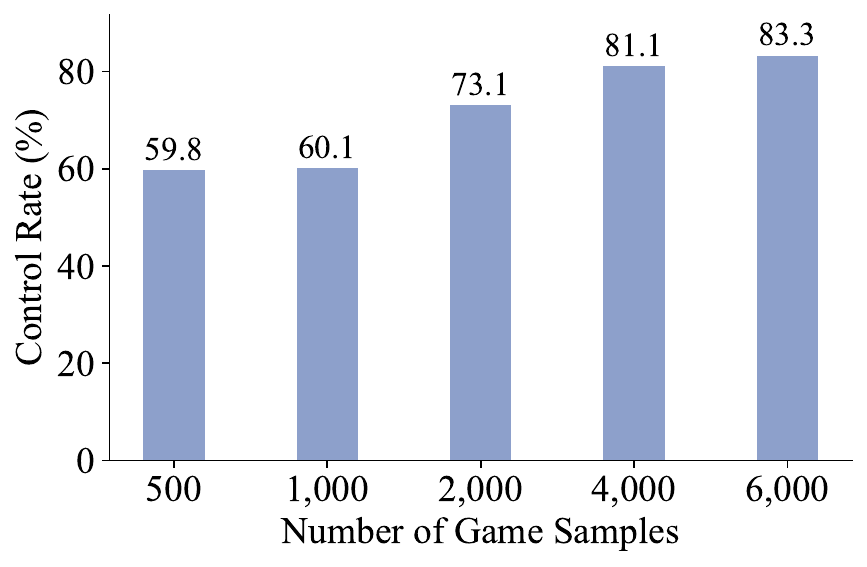}
        \vspace{-3mm}
        \caption{Increasing the number of labeled game samples consistently improves control success on real-world entities.}
        \vspace{-6mm}
        \label{fig:ratio}
    \end{minipage}
\end{figure}

\vspace{-1mm}
\noindent\textbf{Cross-Entity Training Improves Control Success for Individual Entities.} In Table~\ref{tab:training-data}, we evaluate the effect of cross-entity training by training our model on game data combined with real-world data from various entities—bicycles, vehicles, and pedestrians—and testing specifically on bicycles. We make three key observations: (1) RealPlay trained solely on game data fails to control the real-world entity (bicycle); in fact, during generation, the bicycle gradually transforms into a game-style car due to the model's lack of exposure to real-world entity distributions. (2) While labeled game data provides essential supervision, the inclusion of unlabeled real-world data plays a critical role by supplying realistic motion patterns as well as the appearance of entities. As a result, training RealPlay on a combination of game data and bicycle-specific real-world data significantly boosts the control rate to 72.5\%. (3) The additional involvement of other real-world cross-entity data (i.e., vehicles and pedestrians) during training enhances the control of bicycles. This improvement arises because cross-entity data introduces similar motion patterns and shares underlying motion dynamics.

\vspace{-0.5mm}
\noindent\textbf{Real-World Environments Are Significantly Harder to Fit Than Virtual Games.} As a chunk-wise generation model, RealPlay generates each video chunk conditioned on the previously generated chunk and an action signal. This design introduces an error accumulation issue, as the model conditions on its own generated outputs rather than ground-truth data. We analyze how this affects long-term generation in both the game and real-world settings. Figure~\ref{fig:real_game} illustrates the degradation of image quality as the number of generated video chunks increases. 

\vspace{-1mm}
\noindent\textbf{The Number of Video Latents per Chunk Affects Visual Quality.} At each iteration, RealPlay generates a video chunk composed of 4 video latents (equivalent to 16 frames). To enhance user experience with low control latency, it is desirable to minimize the number of frames generated per iteration. However, we observe that generating too few video latents—such as 1 latent (4 frames) per iteration—leads to a noticeable drop in visual quality, as illustrated in Figure~\ref{fig:chunk_x}. This degradation arises because RealPlay leverages a large-scale pre-trained video generation model, CogVideoX, which is originally trained to generate 13 video latents in a single forward pass. Adapting such a model to extremely low-latency, short-horizon generation (e.g. 1 video latent) disrupts its learned temporal coherence and internal consistency, resulting in lower-quality outputs.

\vspace{-1mm}
\noindent\textbf{Co-Improvement of Visual Quality and Control Rate During Training.} Figure~\ref{fig:iteration} illustrates how both the control success rate and Elo scores—which jointly reflect control performance and visual quality based on human evaluation—consistently improve as training progresses.

\vspace{-1mm}
\noindent\textbf{More Labeled Game Samples Improve Control Transfer.} We investigate how the amount of labeled game data affects stable control transfer. Specifically, we use a fixed subset of the real-world training set containing 4K samples, and vary the amount of labeled game data. The control success rate for each setting is reported in Figure~\ref{fig:ratio}.

\vspace{-4mm}
\section{Conclusion}
\vspace{-3mm}
\label{sec:conclusion}
We present RealPlay, a neural network–based real-world game engine that enables interactive, photorealistic video generation conditioned on user control signals. Unlike traditional game engines limited by handcrafted graphics and physics rules, RealPlay leverages a chunk-wise generation framework to support iterative interaction with strong temporal consistency. Through a mixed training paradigm—combining labeled game data with unlabeled real-world videos—RealPlay achieves effective control transfer from virtual to real-world settings. Remarkably, it generalizes to diverse real-world entities, despite being trained only with car-based game supervision. Our results demonstrate RealPlay's potential to bridge simulation and reality, marking a first step toward neural game engines that learn to simulate the real world from data.

\clearpage
\appendix

\section{More Implementation Details}

\noindent\textbf{More Training Details.} Both training stages share the same optimization settings: we use the AdamW optimizer with a learning rate of 1e-5 and a weight decay of 1e-3. A DDPM-style diffusion loss is employed with the default noise scheduler, and training is conducted on 8 NVIDIA A100 (40G) GPUs. The chunk-wise generation model (Stage 1) is trained for 10K iterations, while the final model (Stage 2) is fine-tuned for 15K iterations. During inference, we adopt DDIM with 50 denoising steps.

\noindent\textbf{LAPA.} In Table 1 of the main paper, we present a baseline named CogVideoX-5B (LAPA\cite{ye2024latent}). LAPA is an unsupervised model that learns a pseudo-action $\boldsymbol{a}'$ to enable the transition $\boldsymbol{C}_1 + \boldsymbol{a}' \rightarrow \boldsymbol{C}_2$. The definitions of $\boldsymbol{C}_1$ and $\boldsymbol{C}_2$ are provided in Section 4.1. LAPA adopts a VQ-VAE~\cite{van2017neural} architecture, where the objective is to reconstruct $\boldsymbol{C}_2$ given $\boldsymbol{C}_1$ and a discrete action dictionary that facilitates the reconstruction. In our implementation, we set the dictionary size to 3. We first fine-tune LAPA on our real-world dataset, and then use it to generate a pseudo-action label $\boldsymbol{a}'$ for each training sample. These pseudo-labeled samples are then used to finetune the chunk-wise generation model described in Section~3.1.

\noindent\textbf{Motion Amplitude Computation.} To compute motion amplitude, we first use SAM-2~\cite{ravi2024sam} to segment the entities in each training video, and then apply RAFT~\cite{teed2020raft} to estimate the optical flow. The motion amplitude is defined as the average optical flow magnitude across all training videos for a given entity.

\section{More Experiments}
Table~\ref{tab:control-strategy} compares three action signal injection strategies: (1) the default adaptive LayerNorm described in Section 3.2; (2) self-attention, where the action signal is treated as a token and appended to the input sequence after the language tokens; and (3) cross-attention, where a cross-attention layer is added after the self-attention layer in each attention block, using the output of the self-attention layer as the query and the action token as the key.

\begin{table}[htbp]
\centering
\small
\caption{Comparison of various action signal injection strategies.}
\begin{tabular}{l|cccc|c}
\toprule
\multirow{2}{*}{Strategy} & \multicolumn{4}{c|}{Visual Quality} & Control\\ 
 & Motion & Aesthetic & Imaging  & Dynamic & Success Rate (\%)\\ 
   \midrule
  Self-Attention & 97.7 & 47.3 & 69.2 & 97.5 & 78.3 \\
  Cross-Attention & 97.6 & 47.4 & 68.8 & 100.0 & 77.3 \\
  Adaptive LayerNorm (Default) & 97.5 & 47.8 & 68.9 & 100.0 & 90.0  \\
\bottomrule
\end{tabular}
\label{tab:control-strategy}
\end{table}

\section{More Visualizations}

\begin{figure}[h]
\centering
\includegraphics[width=0.95\textwidth]{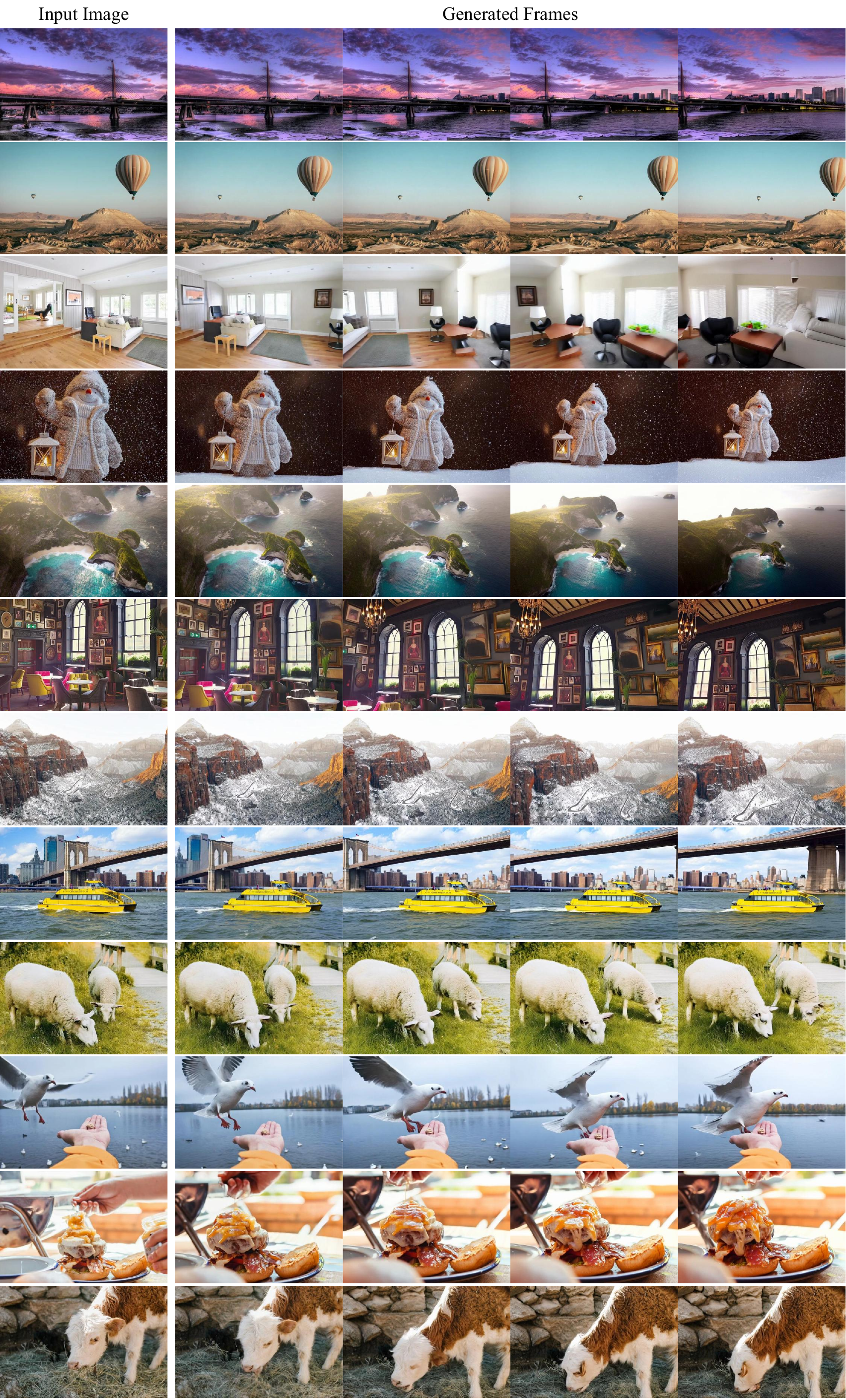}
\caption{Visualizations of videos generated by the chunk-wise generation model.}
\label{fig:stage-1}
\end{figure}

\noindent\textbf{Visualizations of Videos Generated by the Chunk-Wise Generation Model.} 
Section~3.1 details how a pre-trained image-to-video generator is adapted into a chunk-wise generation model. Trained on a general-domain dataset, this model is capable of generating diverse videos in a chunk-wise manner. Figure\ref{fig:stage-1} showcases several examples produced by our chunk-wise generation model.

\begin{figure}[h]
\centering
\includegraphics[width=\textwidth]{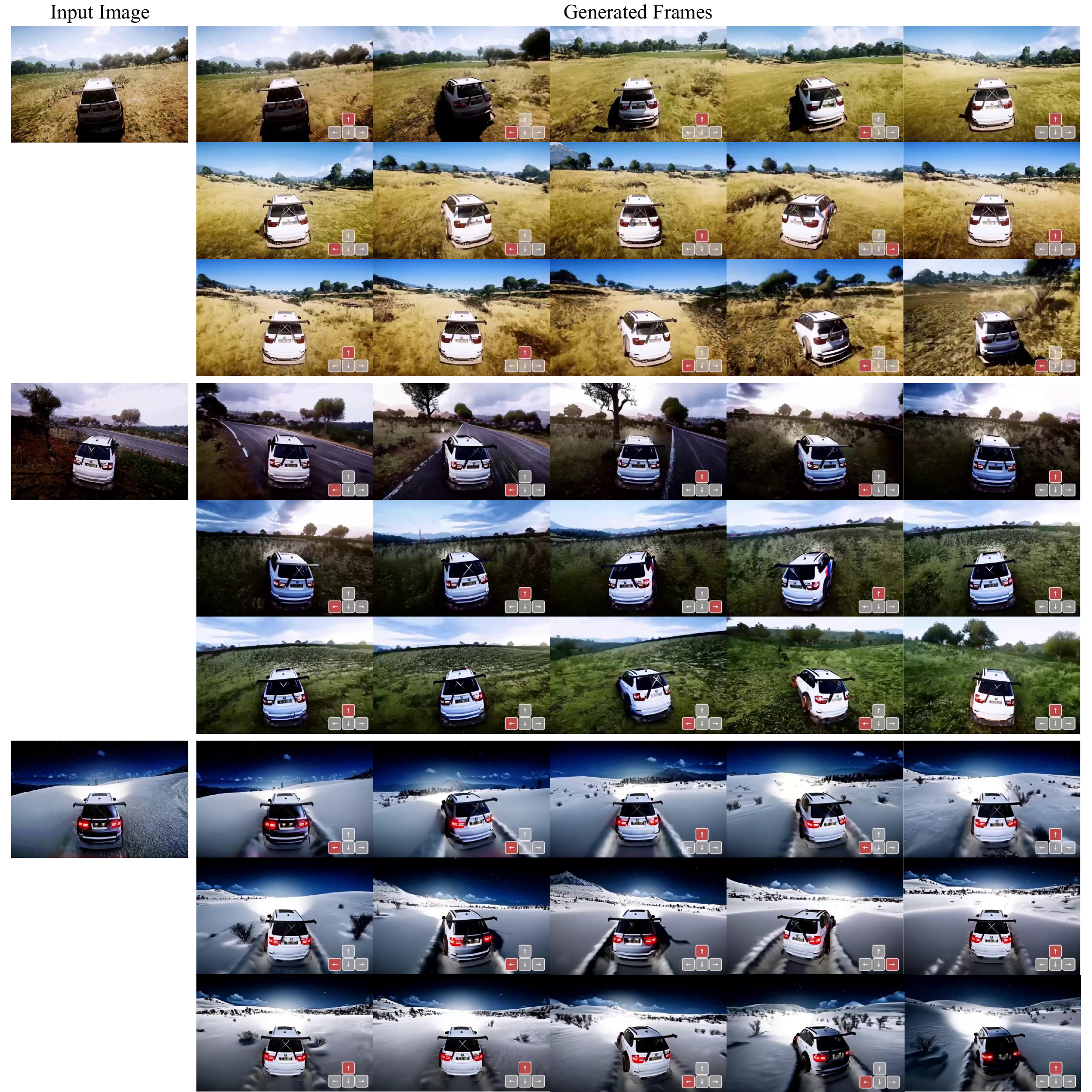}
\caption{Visualization of RealPlay's capability to control cars in diverse game environments.}
\label{fig:stage-2-game}
\end{figure}

\begin{figure}[h]
\centering
\includegraphics[width=0.95\textwidth]{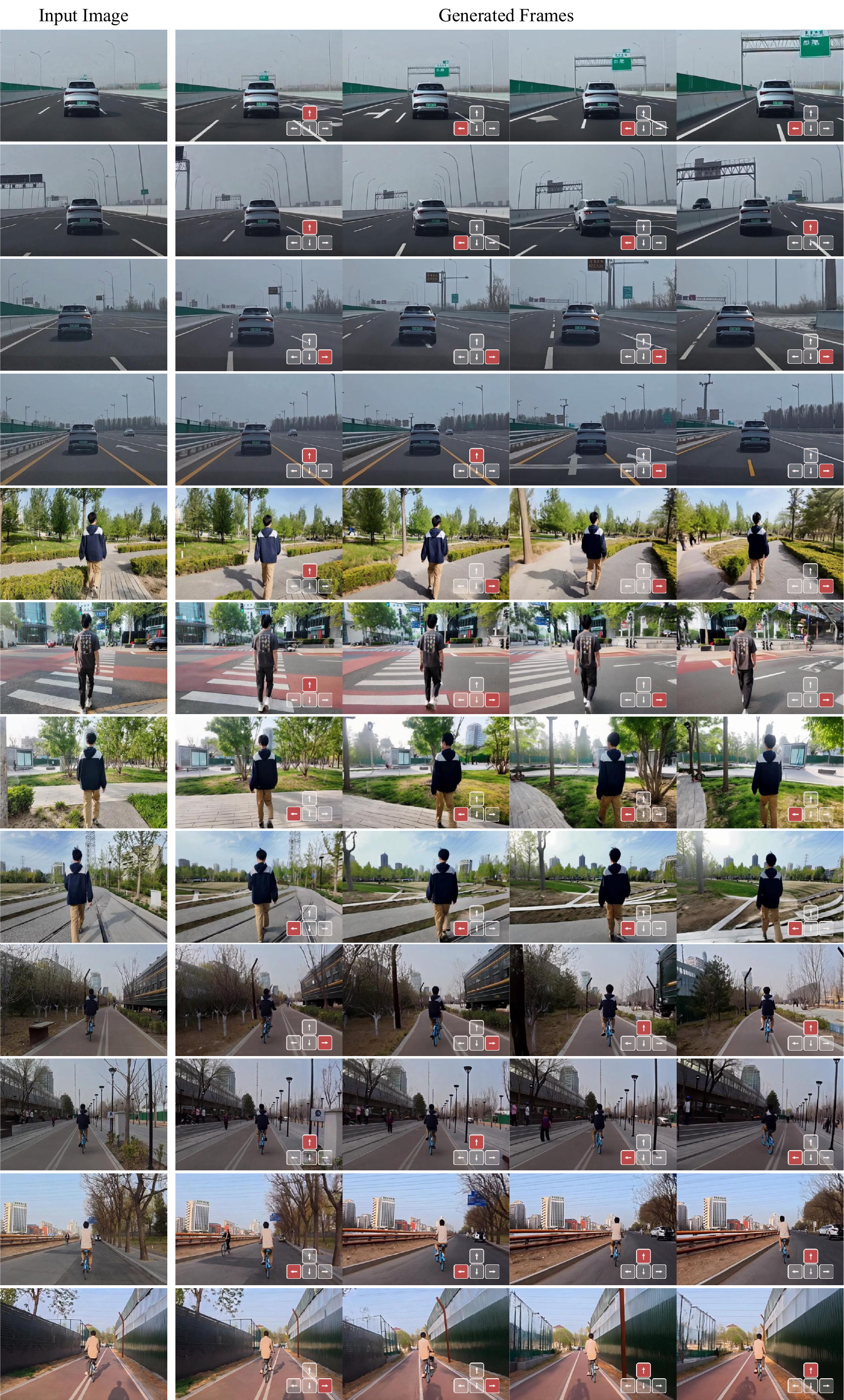}
\caption{RealPlay's effectiveness in controlling diverse real-world entities.}
\label{fig:stage-2-real}
\end{figure}

\begin{figure}[h]
\centering
\includegraphics[width=\textwidth]{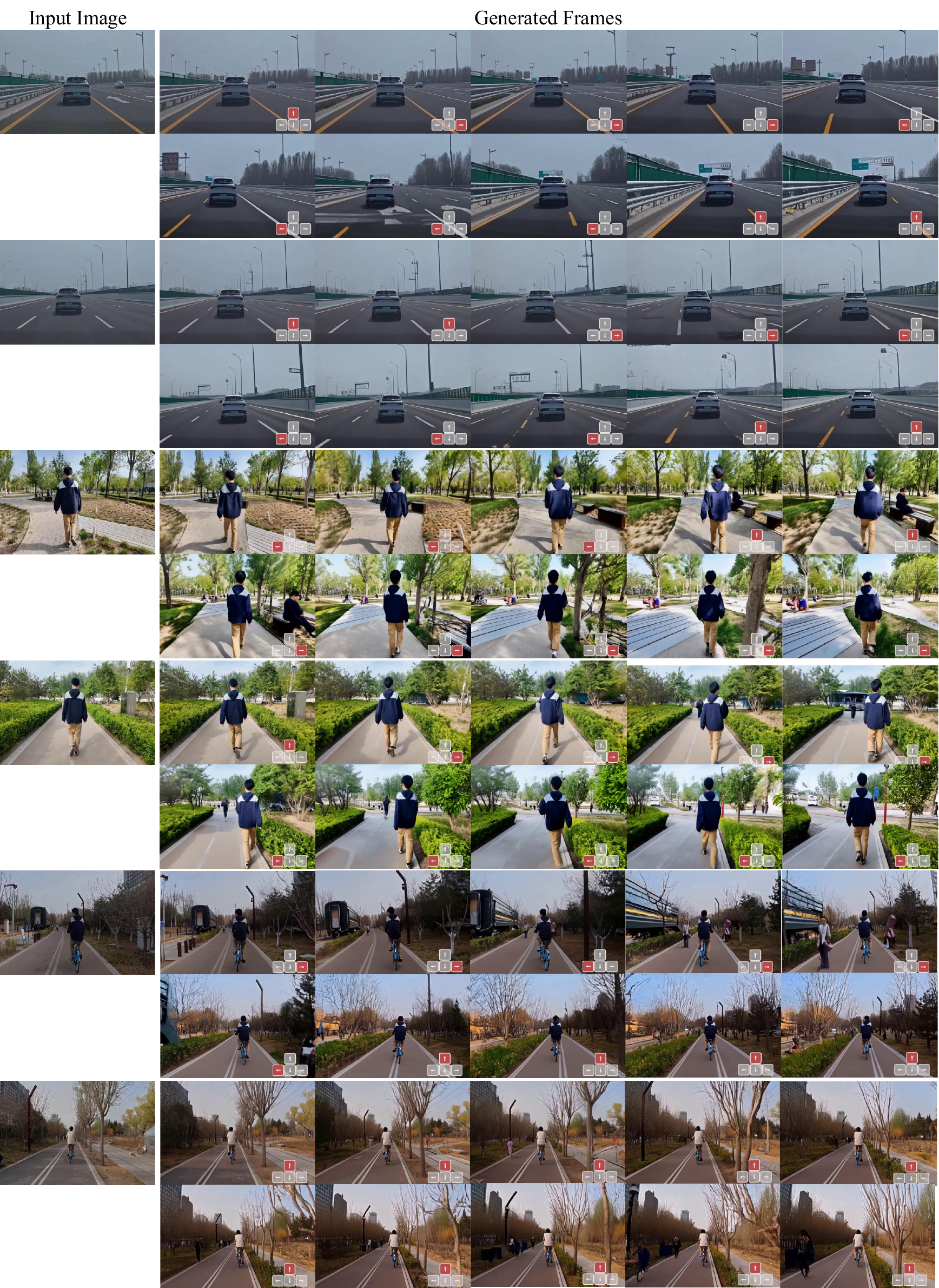}
\caption{Long-horizon videos generated by RealPlay demonstrating sustained control over real-world entities.}
\label{fig:stage-2-long}
\end{figure}

\noindent\textbf{Visualizations of Videos Generated by RealPlay.}
RealPlay is trained on a combination of a labeled car-racing game dataset and an unlabeled real-world dataset containing three types of entities: bicycles, pedestrians, and vehicles. This setup enables control transfer from the game environment to the real world. Figure~\ref{fig:stage-2-game} demonstrates RealPlay’s ability to control cars across diverse game scenes, while Figure~\ref{fig:stage-2-real} illustrates its effectiveness in controlling various real-world entities. Additionally, Figure~\ref{fig:stage-2-long} presents six long-horizon videos generated by RealPlay, demonstrating its capability in sustaining control over real-world entities.

\begin{figure}[h]
\centering
\includegraphics[width=\textwidth]{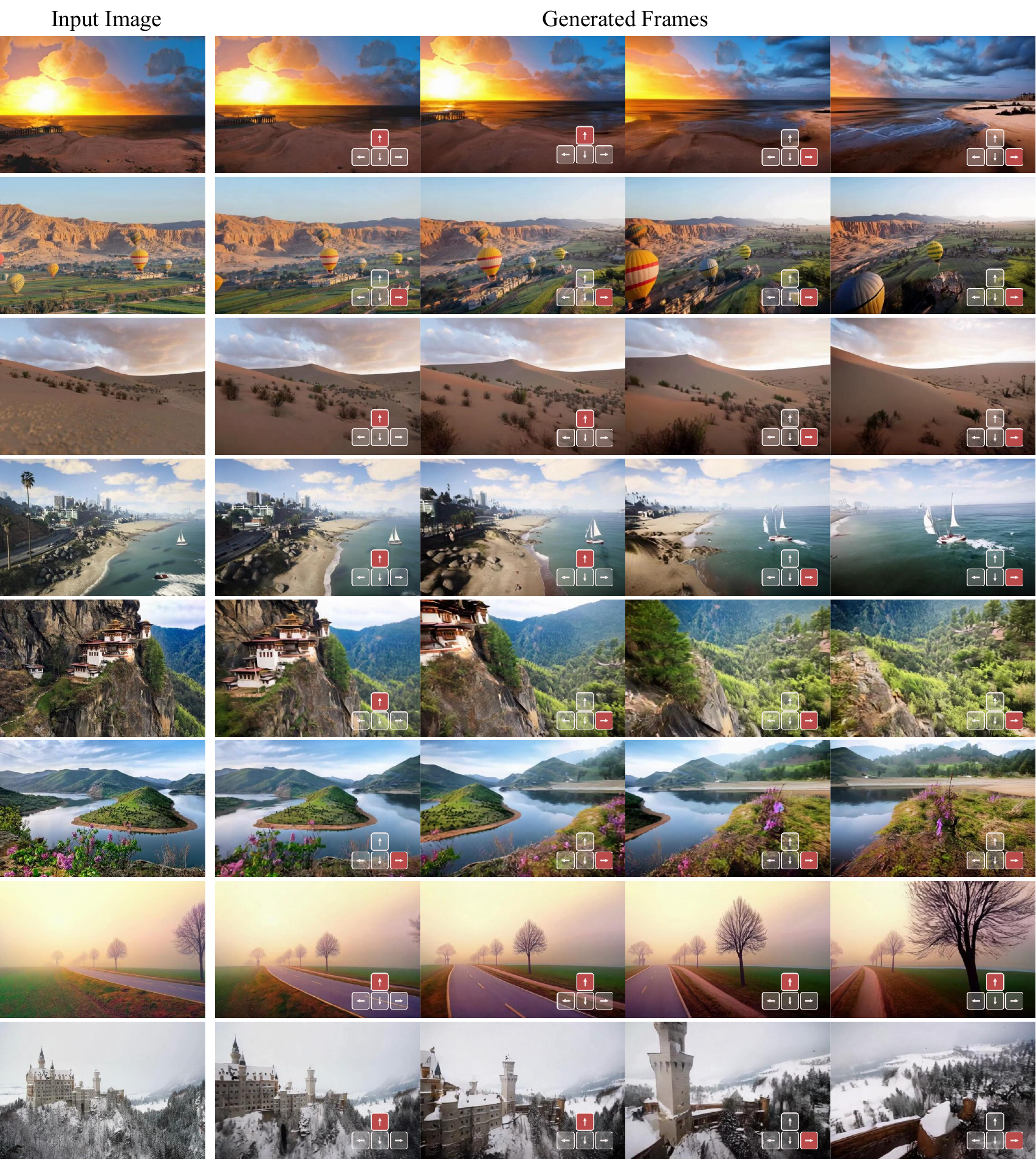}
\caption{RealPlay learns both entity and camera control; when no focused entity is present, it controls the camera alone.}
\label{fig:ood}
\end{figure}

\noindent\textbf{What Happens When Control Signals Are Applied to Videos Without a Clear Entity Focus?} RealPlay enables control over real-world entities to move forward, turn left, and turn right. In typical videos, the target entity is centrally positioned within the frame. Notably, when applying these control signals, the camera is not static—instead, it dynamically follows the trajectory of the entity. In Figure~\ref{fig:ood}, we visualize several examples where there is no clear entity focus. Interestingly, we observe that when control signals are applied to these examples, the camera itself moves in accordance with the control direction. This suggests that RealPlay learns to transfer both entity and camera control: when a focused entity is present in the frame, it controls both the entity and the camera; when no focused entity is present, it controls only the camera. In both cases, the results are visually coherent and pleasant.

\begin{figure}[h]
\centering
\includegraphics[width=\textwidth]{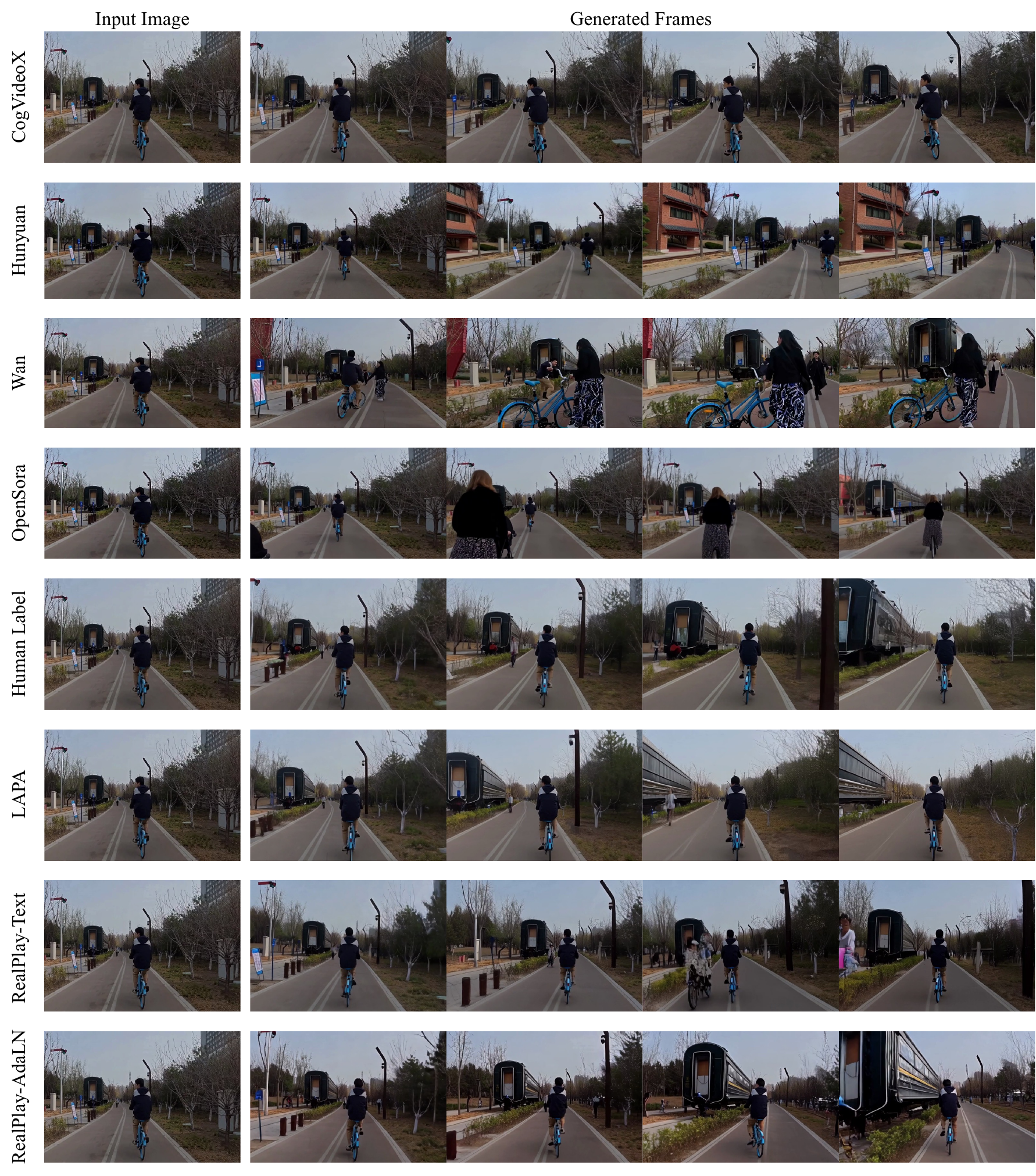}
\caption{Qualitative comparison with baseline methods. The target action is \textbf{move left and then forward}.}
\label{fig:comparison_bicycle}
\end{figure}

\begin{figure}[h]
\centering
\includegraphics[width=\textwidth]{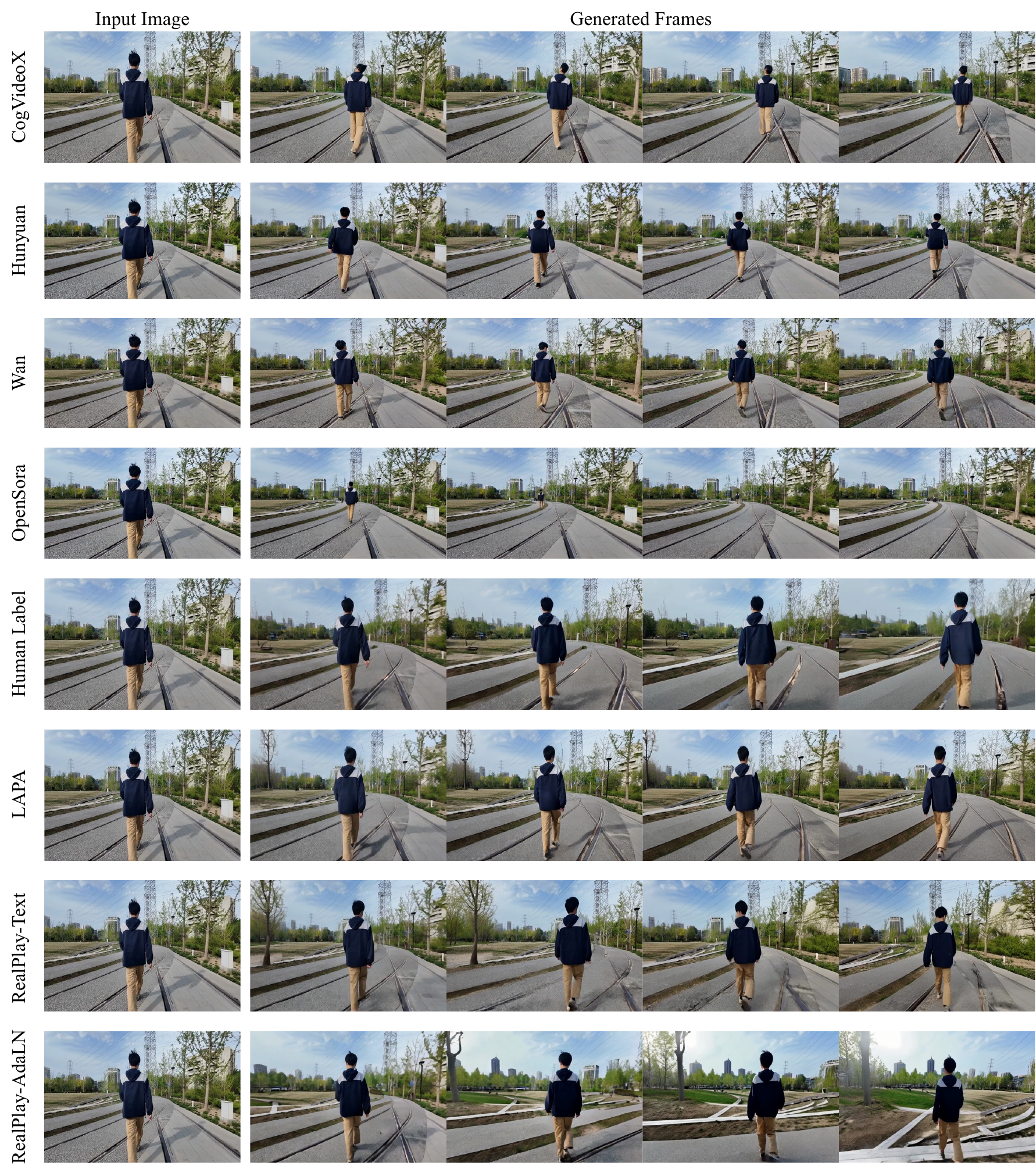}
\caption{Qualitative comparison with baseline methods. The target action is \textbf{move left}.}
\label{fig:comparison_human}
\end{figure}

\noindent\textbf{Qualitative Comparison with Baseline Methods.} We provide a qualitative comparison with baseline methods listed in Table 1 of our main paper. As shown in Figure \ref{fig:comparison_bicycle} and Figure \ref{fig:comparison_human}, in comparison to baselines, our approach achieves more precise control while maintaining superior visual quality.

\section{Limitations and Broader Impacts}
This work represents an initial step toward leveraging video generation techniques to build real-world games, where visuals are photorealistic and the game engine is powered by a data-driven neural network. RealPlay depends on a pre-trained general-purpose video generator—specifically, CogVideoX-5B in this study. Due to the large model size and the relatively high resolution of the generated videos, RealPlay cannot yet operate in real-time. Several techniques could potentially accelerate inference, such as model distillation and few-step denoising methods like the Shortcut~\cite{frans2024one} model. We consider these as promising directions for future optimization and deployment.

\clearpage
{\small
\bibliographystyle{abbrvnat}
\bibliography{ref}
}



\end{document}